\pdfoutput=1

\documentclass[11pt]{article}

\usepackage{acl}

\usepackage{times}
\usepackage{latexsym}

\usepackage[T1]{fontenc}

\usepackage[utf8]{inputenc}

\usepackage{microtype}
\usepackage{graphicx}
\usepackage{multirow}
\usepackage{enumitem}
\usepackage{booktabs}
\usepackage{listings}

\makeatletter
    \newcommand\crulefill{\leavevmode
    \begingroup 
    \setlength{\dimen@}{0.5ex}
    \addtolength{\dimen@}{0.4pt}
    \leaders\hrule height\dimen@ depth -0.5ex \hfill
    \endgroup
    \kern\z@}

\title{EventGraph: Event Extraction as Semantic Graph Parsing}

\author{Huiling You,$^1$ David Samuel,$^1$ Samia Touileb,$^2$ \and Lilja Øvrelid$^1$ \\
         $^1$University of Oslo\\
         $^2$University of Bergen \\ 
         \texttt{\{huiliny, davisamu, liljao\}@ifi.uio.no} \\
         \texttt{samia.touileb@uib.no}
}

\begin{document}
\maketitle
\begin{abstract}
Event extraction involves the detection and extraction of both the event triggers and corresponding event arguments. Existing systems often decompose event extraction into multiple subtasks, without considering their possible interactions. In this paper, we propose EventGraph, a joint framework for event extraction, which encodes events as graphs. We represent event triggers and arguments as nodes in a semantic graph. Event extraction therefore becomes a graph parsing problem, which provides the following advantages: 1) performing event detection and argument extraction jointly; 2) detecting and extracting multiple events from a piece of text; and 3) capturing the complicated interaction between event arguments and triggers. Experimental results on ACE2005 show that our model is competitive to state-of-the-art systems and has substantially improved the results on argument extraction. Additionally, we create two new datasets from ACE2005 where we keep the entire text spans for event arguments, instead of just the head word(s). Our code and models are released as open-source.\footnote{\url{https://github.com/huiling-y/EventGraph}}
\end{abstract}

\section{Introduction}

\begin{figure}[t!]
\centering
\includegraphics[width=\columnwidth]{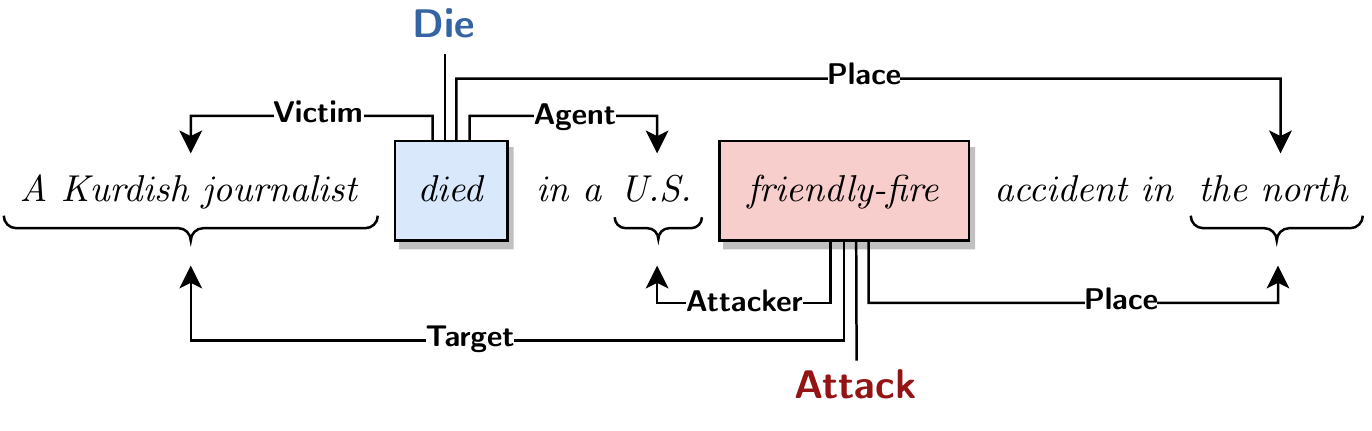}
\caption{\label{eg_event} Example of an \texttt{Attack} and a \texttt{Die} events in the sentence ``\textit{A Kurdish journalist died in a U.S. friendly-fire accident in the north}.'' }
\end{figure}

Event extraction aims at extracting event-related information from unstructured texts into structured form (i.e. triggers and arguments), according to a predefined event ontology \citep{ahn-2006-stages, doddington-etal-2004-automatic}. In these types of ontologies, events are characterized by event triggers, and comprise a set of predefined argument types. Figure \ref{eg_event} shows an example of a sentence containing two events, an \texttt{Attack} event triggered by \textit{``friendly-fire''} and a \texttt{Die} event triggered by \textit{``died''}; the two events share the same arguments, but each plays a different role in the specific event. For instance, \textit{``U.S.''} is the \texttt{Agent} in the \texttt{Die} event, but plays the role of \texttt{Attacker} in the \texttt{Attack} event.

As opposed to dividing event extraction into independent subtasks, we take advantage of recent advances in semantic dependency parsing \cite{dozat-manning-2018-simpler, samuel-straka-2020-ufal} and develop an end-to-end event graph parser, dubbed EventGraph. We adopt intuitive graph encoding to represent the event mentions of a piece of text in a single event graph, and directly generate these event graphs from raw texts. We evaluate our EventGraph system on ACE2005 (LDC2006T06).\footnote{\url{https://catalog.ldc.upenn.edu/LDC2006T06}} Our model achieves competitive results with state-of-the-art models, and substantially improves the results on event argument extraction. The main contributions of this work are:

\begin{enumerate}
    \item We propose EventGraph, a text-to-event framework that solves event extraction as semantic graph parsing. The model does not rely on any language-specific features or event-specific ontology, so it can easily be applied to new languages and new datasets.
    \item We design an intuitive graph encoding approach to represent event structures in a single event graph.
    \item The versatility of our approach allows for an effortless decoding of full trigger and argument mentions. We create two novel and more challenging datasets from ACE2005, and provide corresponding benchmark results.
\end{enumerate}

\section{Related work}

Our work is closely related to two research directions, event extraction and semantic parsing.

Supervised event extraction is an established research area in NLP. There are different methods to obtain the structured information of an event, and the mainstream methods can be divided into: 1) classification-based methods: treat event extraction as several classification subtasks, and either solve them separately in a pipeline-based manner \citep{ji-grishman-2008-refining, li-etal-2013-joint, liu-etal-2020-event, du-cardie-2020-event, li-etal-2020-event} or jointly infer multiple subtasks \citep{yang-mitchell-2016-joint, nguyen-etal-2016-joint-event, liu-etal-2018-jointly, wadden-etal-2019-entity, lin-etal-2020-joint}; 2) generation-based approaches: formulate event extraction as a sequence generation problem \citep{paolini2021structured,lu-etal-2021-text2event, li-etal-2021-document, hsu2022degree}; 3) prompt tuning methods: inspired by natural language understanding tasks, these approaches take advantage of ``discrete prompts'' \citep{shin-etal-2020-autoprompt,gao-etal-2021-making,li-liang-2021-prefix,liu-etal-2022-dynamic}.

Meaning Representation Parsing has seen significant interest in recent years  \citep{oepen2014semeval,oepen2015semeval,oepen-etal-2020-mrp}. Unlike syntactic dependency representations, these semantic representations are crucially not trees, but rather general graphs, characterised by potentially having multiple entry points (\textit{roots}) and not necessarily being connected, since not every token is a node in the graph.
There has further been considerable progress in developing variants of both transition-based and graph-based dependency parsers capable of producing such semantic graphs \cite{hershcovich-etal-2017-transition, dozat-manning-2018-simpler,samuel-straka-2020-ufal}. 

A recent and highly relevant development in the current context has been the application of semantic parsers to NLP tasks beyond meaning representation parsing. These approaches rely on the reformulation of task-specific representations to semantic dependency graphs. For example, \citet{yu2020named} exploit the parser of \citet{dozat-manning-2018-simpler} to predict spans of named entities, while \citet{kurtz-etal-2020-end} phrase the task of negation resolution \cite{morante2012conandoyleneg} as a graph parsing task with promising results. Recently, \newcite{barnes-etal-2021-structured} proposed a dependency parsing approach to extract opinion tuples from text, dubbed Structured Sentiment Analysis, and a recent shared task dedicated to this task demonstrated the usefulness of graph parsing approaches to sentiment analysis \cite{barnes2022semeval}. Most similar to our work is the work by \citet{samuel-etal-2022-direct} which adapts the PERIN parser \cite{samuel-straka-2020-ufal} to parse directly from raw text into sentiment graphs.

\section{Event graph representations}

\begin{figure}[t!]
\centering
\includegraphics[width=\columnwidth]{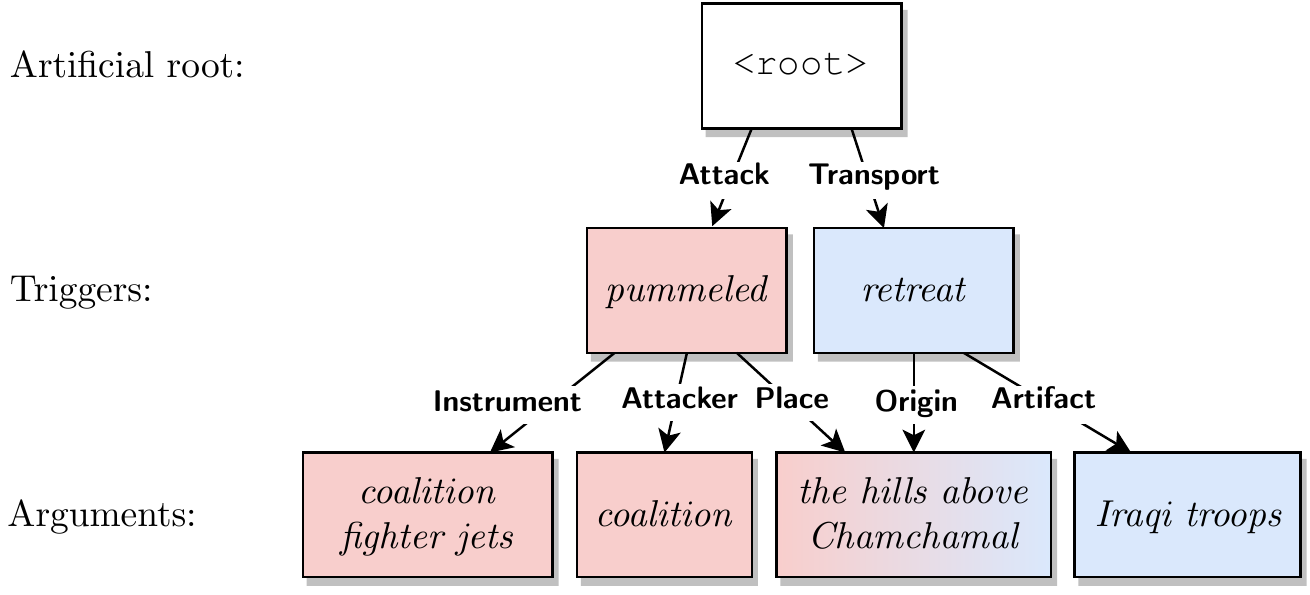}
\caption{\label{event_graph} Event graph for the sentence  \textit{``That's because coalition fighter jets pummeled this Iraqi position on the hills above Chamchamal and Iraqi troops made a hasty retreat.''}}
\end{figure}

We adopt an efficient ``labeled-edge'' representation for event graph encoding within the scope of a sentence. Each node in an event graph corresponds to either an event trigger or an argument, which is anchored to a unique text span in a sentence, except for the top node, which is only a dummy node for every event graph. The edges are constrained only between the top node and an event trigger, or between an event trigger and an argument, with the corresponding edge label as an event type or argument role. The ``labeled-edge'' encoding has the ability to represent: 1) multiple event mentions; 2) nested structures (overlapping between arguments or trigger-argument); 3) multiple argument roles of a single argument. Taking the event graph from Figure \ref{event_graph} as example, the sentence contains two event mentions, which share the same argument \textit{``the hills above Chamchamal''} but as different roles, and the argument \textit{``coalition''} is nested inside the argument \textit{``coalition fighter jets''}.

\section{Event parsing} 

\begin{figure}[t!]
\centering
\includegraphics[width=\columnwidth]{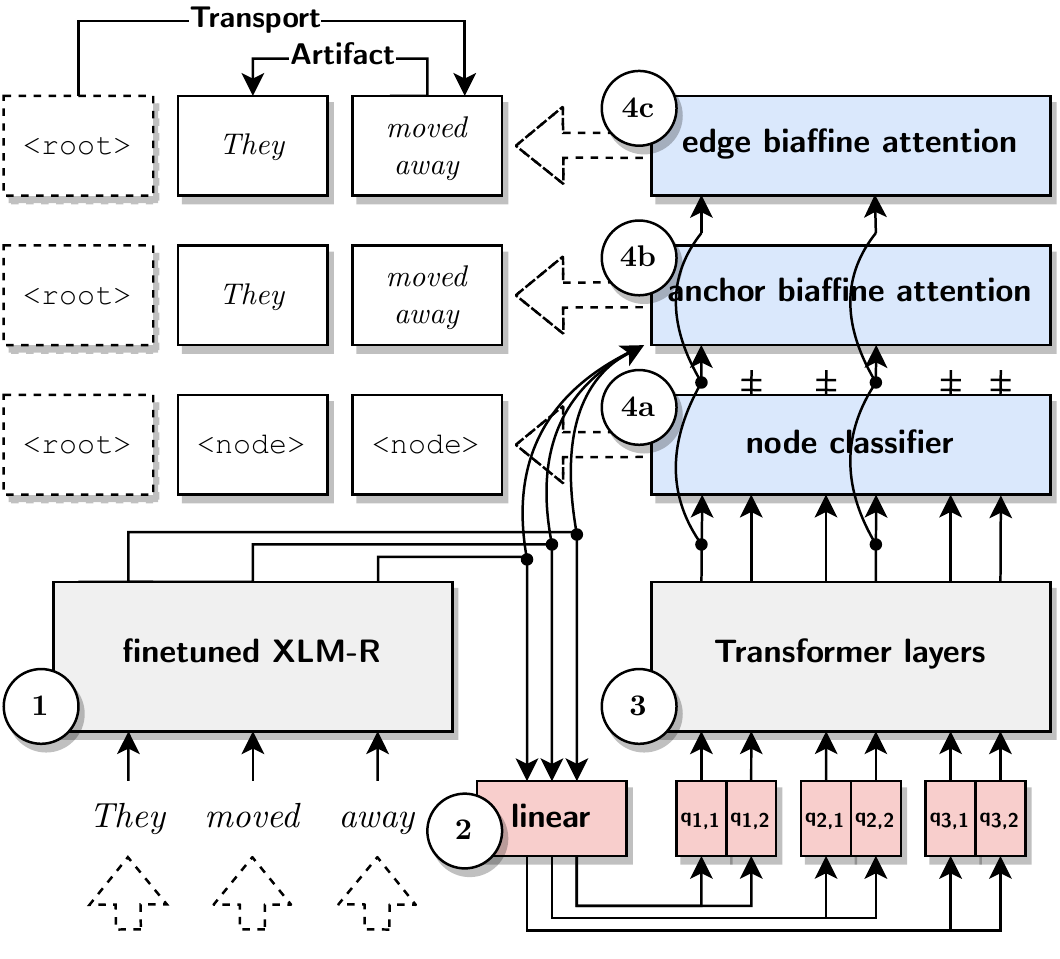}
\caption{\label{parser} EventGraph architecture. 1) the input gets a contextualized representation, 2) queries are generated for every input token, 3) queries are further processed with a decoder to predict 4a) node presence, 4b) node anchors, and 4c) edge labels.}
\label{fig:perin}
\end{figure}

EventGraph is an adaptation of PERIN \citep{samuel-straka-2020-ufal}, a general permutation-invariant framework for text-to-graph parsing. Given the ``labeled-edge'' encoding for event graphs, we create EventGraph by customizing the modules of PERIN as illustrated in Figure \ref{fig:perin}, which contains three classifiers to generate nodes, anchors, and edges, respectively. Each input sequence is processed by four modules of EventGraph to generate a final structured representation.

\paragraph{Encoder} We use the large version of XLM-R \citep{conneau-etal-2020-unsupervised} as the encoder to obtain contextualized representations of the input sequence; each token gets a contextual embedding via a learned subword attention layer over the subwords.

\paragraph{Query generator} We use a linear transformation layer to map each embebbed token onto $n$ queries.

\paragraph{Decoder} The decoder is a stack of Transformer encoder layers \citep{vaswani2017attention} without positional encoding, which is permutation-invariant (non-autoregressive); the decoder processes and augments the queries of each token by modelling the inter-dependencies between queries.

\paragraph{Parser head}  It consists of three classifiers: 
a) the \textbf{node classifier} is a linear classifier that predicts node presence by classifying the augmented queries of each token; since more than one query is generated for each token, a single token can produce more than one node; 
b) the \textbf{anchor biaffine classifier} \citep{dozat2017deep} uses deep biaffine attention between the augmented queries and contextual embeddings of each token to map the predicted nodes to surface tokens; 
c) the \textbf{edge biaffine classifier} uses two deep biaffine attention modules to process generated nodes and predict edge presence between a pair of nodes and the edge label.

Given a piece of text, EventGraph generates its corresponding graph, and it is effortless to extract the structured information of event mentions from the nodes and edges.\footnote{The tool for conversion between event mentions and event graphs is included in our codes.}

\section{Experimental setup}

\subsection{Datasets}

We evaluate our system on the widely used benchmark dataset  ACE2005\footnote{\url{https://catalog.ldc.upenn.edu/LDC2006T06}} (LDC2006T06). The ACE2005 dataset contains 599 English documents annotated for several tasks, entities, values, relations, and events, with an event ontology of 33 event types, and 35 argument roles. Event arguments come from both entities and values. The annotation of an entity also includes its head word(s); for instance, from Table \ref{tab:eg_event}, entity ``the Iraqi government's key center of power'' has ``center'' as its head word. Following previous works \citep{wadden-etal-2019-entity, lin-etal-2020-joint, wang-etal-2019-hmeae}, we preprocess the dataset (details in Appendix \ref{sec:preprocessing}) and obtain the following configurations:

\begin{enumerate}
    \item \textbf{ACE05-E}: \citet{wadden-etal-2019-entity} keep 22 event argument roles (excluding ``time'' and ``value'' event arguments), ignore events with multi-token trigger(s), and use only the head word(s) of event arguments.
    \item \textbf{ACE05-E${^+}$}: similar to \citet{wadden-etal-2019-entity}, \citet{lin-etal-2020-joint} only use 22 event argument roles and keep only the head word(s) of event arguments, but keep events with multi-token trigger(s).
    \item \textbf{ACE05-E$^{++}$}: we create a new dataset that keeps the full text spans for event triggers and event arguments, but also keep 22 argument roles for comparing with previous work.
    \item \textbf{ACE05-E$^{+++}$}: we create another dataset that keeps all the 35 argument roles in ACE2005, with full text spans for event triggers and arguments. 
\end{enumerate}

Table \ref{tab:eg_event} shows how an event mention is extracted in ACE05-E${^+}$ and ACE05-E$^{++}$, and the same event is not present in ACE05-E. Although keeping the full text spans for arguments makes the task of argument extraction more difficult, we believe that the extracted events are more informative and self-contained.

\begin{table}[t!]
\resizebox{\columnwidth}{!}{%
\begin{tabular}{@{}lll@{}}
\toprule
& \textbf{ACE05-E${^+}$} & \textbf{ACE05-E$^{++}$} \\
\midrule
\texttt{Trigger} & \textit{``push ahead''} & \textit{``push ahead''} \\ 
\texttt{Destination} & \textit{``center''} & \textit{``the Iraqi government's key center of power''} \\
\texttt{Artifact} & \textit{``forces''} & \textit{``American forces''} \\
\bottomrule
\end{tabular} %
}
\caption{\label{tab:eg_event} A \texttt{Transport} event in ``\textit{Well, as American forces do push ahead toward the Iraqi government's key center of power, British forces are keeping up their work to the south of the Iraqi capital}'', and corresponding extracted events in ACE05-E${^+}$ and ACE05-E$^{++}$.}
\end{table}

\begin{table}[t!]
\resizebox{\columnwidth}{!}{%
\begin{tabular}{@{}ll@{\hspace{2em}}rrr@{} }
\toprule 
\textbf{Dataset} & \textbf{Split} & \textbf{\# Sentences} & \textbf{\# Events} & \textbf{\# Roles} \\
\midrule

\multirow{3}{*}{ACE05-E} & Train & 17\,172 & 4\,202 & 4\,859 \\
& Dev & 923 & 450 & 605 \\
& Test & 832 & 403 & 576 \\ \midrule
\multirow{3}{*}{ACE05-E${^+}$} & Train & 19\,216 & 4\,419 & 6\,607 \\
& Dev & 901 & 468 & 759 \\
& Test & 676 & 424 & 689 \\ \midrule
\multirow{3}{*}{ACE05-E${^{++}}$} & Train & 15\,603 & 4\,416 & 6\,513 \\
& Dev & 893 & 509 & 802 \\
& Test & 729 & 424 & 685 \\ \midrule
\multirow{3}{*}{ACE05-E${^{+++}}$} & Train & 15\,603 & 4\,416 & 7\,844 \\
& Dev & 893 & 509 & 945 \\
& Test & 729 & 424 & 894 \\
\bottomrule

\end{tabular}%
}
\caption{\label{tab:data-stats} Statistics of the preprocessed ACE2005 datasets.}
\end{table}

\begin{table}[t!]
\resizebox{\columnwidth}{!}{%
\begin{tabular}{@{}lr@{\hspace{2em}}rrr@{} }
\toprule 
\multirow{2}{*}{\textbf{Dataset}} & \textbf{Triggers} & \multicolumn{3}{@{}c@{}}{\crulefill\,\,\,\textbf{Arguments}\,\,\,\crulefill} \\
 & Avg. Len & Avg. Len & Single-token  & Multi-token \\
\midrule

ACE05-E & 1.00 & 1.18 & 86.2\% & 13.8\% \\ 
ACE05-E${^+}$ & 1.06 & 1.17 & 88.0\% & 12.0\% \\
ACE05-E${^{++}}$ & 1.05 & 2.86 & 43.5\% & 56.5\% \\
ACE05-E${^{+++}}$ & 1.05 & 2.82 & 43.2\% & 56.8\% \\

\bottomrule

\end{tabular}%
}
\caption{\label{tab:data-stats2} Statistics of event triggers and arguments. We report the average lengths of triggers and arguments; for arguments, we also report the percentages of single-token and multi-token arguments.}
\end{table}

\subsection{Evaluation metric}

We report Precision \emph{P}, Recall \emph{R}, and \emph{F1} scores for each of the following evaluation criteria \citep{wadden-etal-2019-entity,lin-etal-2020-joint}:

\begin{itemize}
    \item \textbf{Trigger classification (Trg-C)}: an event trigger is correctly predicted if its offsets and event type matches the gold trigger.
    \item \textbf{Argument classification (Arg-C)}: an event argument is correctly predicted if its offsets, argument role, and event type match the gold argument.
\end{itemize}

For argument classification, in order to have a better insight into our models' performance on multi-token arguments, we include another metric based on token-level span overlap for argument identification, instead of perfect match.
\begin{itemize}
    \item \textbf{Token-level span overlap}: an event argument is correctly identified if its offsets have 80\%\footnote{This metric only affects arguments longer than 5 tokens. Arguments containing fewer than 5 tokens are still evaluated with perfect match.} overlap (token-level) with the gold argument, and correctly predicted if its argument role and event type match the gold argument.
\end{itemize}

\subsection{System comparisons}

We compare EventGraph to the following event extraction systems: 
1) DYGIE++ \citep{wadden-etal-2019-entity}: a span-based framework capturing both local and global contexts; 
2) ONEIE \citep{lin-etal-2020-joint}: an end-to-end framework for general information extraction; 
3) TEXT2EVENT \citep{lu-etal-2021-text2event}: a generation-based model for sequence-to-event generation; 
4) GTEE-DYNPREF \citep{liu-etal-2022-dynamic}: a template-based method for text-to-event generation.

\subsection{Implementation details}

Our code is built upon the official implementation of the PERIN parser \citep{samuel-straka-2020-ufal}.\footnote{\url{https://github.com/ufal/perin}} Details about our training setup and hyperparameter settings are given in Appendix \ref{sec:training-details}. For each dataset, we train 5 models with 5 different random seeds, and report the means and standard deviations of the corresponding results.

\section{Results and discussion}

In Table \ref{tab:results}, we compare our results on ACE05-E and ACE05-E${^+}$ with the previous systems. On both datasets, EventGraph achieves SOTA results on Arg-C over all metrics, with an improvement of 7 percentage points on ACE-E and more than 10 percentage points on ACE05-E${^+}$ in \emph{F1} scores. For Trg-C, despite not beating the SOTA systems, our results are still very competitive.

On the two new datasets that we created, EventGraph has achieved overall competitive results (Table~\ref{tab:results}). On ACE-E$^{++}$, despite having longer and more complicated arguments, EventGraph has generated comparable results to those of GTEE-DYNPREF (current SOTA) on ACE-E$^{+}$. On ACE-E$^{+++}$, even though the argument role set is expanded from 22 to 35 argument roles, the results of EventGraph on Arg-C remain stable.

Results show that EventGraph performs well on joint modelling of event triggers and arguments, and benefits from longer text spans for event triggers and arguments. When the full text spans of arguments are used, the model receives more training signals, so it has more information in differentiating sentences containing events from those without, as shown in Table \ref{tab:results3}, and thus identifying event triggers, which is also shown by the increasing Trg-C scores from ACE-E and ACE-E$^{+}$ to ACE-E$^{++}$ and ACE-E$^{+++}$. For instance, as the example in Table \ref{tab:eg_event} shows, ``the Iraqi government's key center of power'' is less ambiguous than mere ``center''. As shown in Table \ref{tab:data-stats2}, the average argument length of ACE-E$^{++}$ and ACE-E$^{+++}$ is much longer, but the average trigger length is very similar across the four datasets; it is also evident that single-token arguments make up a large proportion of all arguments, even for ACE-E$^{++}$ and ACE-E$^{+++}$, so there is a long tail in argument length distribution. For longer arguments, it is more difficult to obtain a perfect match with a gold argument, so we observe decreasing Arg-C scores when EventGraph is evaluated on ACE-E$^{++}$ and ACE-E$^{+++}$.

To further look into our model's performance on identifying multi-token event arguments, especially those containing more than 5 tokens, we further report Arg-C scores based on token-level span overlap. As shown in Table \ref{res:args}, when we relax argument identification from perfect match to 80\% token-level span overlap, the scores of Arg-C increase consistently, especially those of ACE-E$^{++}$ and ACE-E$^{+++}$, now comparable to the results on ACE-E and ACE-E$^{+}$.

\begin{table}[t!]
\resizebox{\columnwidth}{!}{%
\begin{tabular}{@{}l@{\hspace{2em}}ccc@{\hspace{2em}}ccc@{} }
\toprule 
\multirow{2}{*}{\textbf{Model}} & \multicolumn{3}{@{}c@{\hspace{2em}}}{\textbf{Triggers (Trg-C)}} & \multicolumn{3}{@{}c@{}}{\textbf{Arguments (Arg-C)}} \\

& P & R & F1 & P & R & F1 \\ \midrule
\multicolumn{7}{@{}c@{}}{\raisebox{0.8ex}{\small\textbf{Dataset: ACE05-E}}} \\ 
DYGIE++ & \crulefill & \crulefill & 69.7\hphantom{$^{\pm0.0}$} & \crulefill & \crulefill & 48.8\hphantom{$^{\pm0.0}$}  \\
ONEIE & \crulefill & \crulefill & \textbf{74.7}\hphantom{$^{\pm0.0}$} & \crulefill & \crulefill & 56.8\hphantom{$^{\pm0.0}$} \\
GTEE-DYNPREF & 63.7\hphantom{$^{\pm0.0}$} & \textbf{84.4}\hphantom{$^{\pm0.0}$} & 72.6\hphantom{$^{\pm0.0}$} & 49.0\hphantom{$^{\pm0.0}$} & 64.8\hphantom{$^{\pm0.0}$} & 55.8\hphantom{$^{\pm0.0}$} \\
\textbf{EventGraph} & \textbf{66.5$^{\pm 0.7}$} & 71.0$^{\pm 0.9}$ & 68.6$^{\pm 0.7}$ & \textbf{63.4$^{\pm 2.7}$} & \textbf{67.3$^{\pm 2.0}$} & \textbf{65.3$^{\pm 2.2}$} \\ \midrule

\multicolumn{7}{@{}c@{}}{\raisebox{0.8ex}{\small\textbf{Dataset: ACE05-E$^{+}$}}} \\
ONEIE & \textbf{72.1}\hphantom{$^{\pm0.0}$} & 73.6\hphantom{$^{\pm0.0}$} & 72.8\hphantom{$^{\pm0.0}$} & 55.4\hphantom{$^{\pm0.0}$} & 54.3\hphantom{$^{\pm0.0}$} & 54.8\hphantom{$^{\pm0.0}$} \\
TEXT2EVENT & 71.2\hphantom{$^{\pm0.0}$} & 72.5\hphantom{$^{\pm0.0}$} & 71.8\hphantom{$^{\pm0.0}$} & 54.0\hphantom{$^{\pm0.0}$} & 54.8\hphantom{$^{\pm0.0}$} & 54.4\hphantom{$^{\pm0.0}$}  \\
GTEE-DYNPREF & 67.3\hphantom{$^{\pm0.0}$} & \textbf{83.0}\hphantom{$^{\pm0.0}$} & \textbf{74.3}\hphantom{$^{\pm0.0}$} & 49.8\hphantom{$^{\pm0.0}$} & 60.7\hphantom{$^{\pm0.0}$} & 54.7\hphantom{$^{\pm0.0}$} \\ 
\textbf{EventGraph} & 70.0$^{\pm 1.1}$ & 70.0$^{\pm 1.2}$ & 70.0$^{\pm 1.1}$ & \textbf{64.5$^{\pm 1.0}$} & \textbf{66.4$^{\pm 2.6}$} & \textbf{65.4$^{\pm 1.7}$} \\ \midrule

\multicolumn{7}{@{}c@{}}{\raisebox{0.8ex}{\small\textbf{Dataset: ACE05-E$^{++}$}}} \\
\textbf{EventGraph} & 72.9$^{\pm 1.3}$ & 75.2$^{\pm 1.9}$ & 74.0$^{\pm 1.5}$ & 57.3$^{\pm 0.8}$ & 59.9$^{\pm 1.2}$ & 58.6$^{\pm 0.9}$ \\ \midrule

\multicolumn{7}{@{}c@{}}{\raisebox{0.8ex}{\small\textbf{Dataset: ACE05-E$^{+++}$}}} \\
\textbf{EventGraph} & 72.4$^{\pm 0.7}$ & 75.9$^{\pm 1.0}$ & 74.0$^{\pm 0.7}$ & 56.9$^{\pm 0.6}$ & 58.2$^{\pm 0.9}$  & 57.5$^{\pm 0.6}$ \\ \bottomrule
\end{tabular}%
}
\caption{\label{res:ace-e} Results on ACE05-E, ACE05-E$^{+}$, ACE05-E$^{++}$, and ACE05-E$^{+++}$. We report the average performance of 5 runs with different random seeds, together with the standard deviations. For clarity, we bold the highest scores.} 
\label{tab:results}
\end{table}

\begin{table}[t!]
\resizebox{\columnwidth}{!}{%
\begin{tabular}{@{}l@{\hspace{2em}}ccc@{\hspace{2em}}ccc@{} }
\toprule 
\multirow{2}{*}{\textbf{Dataset}} & \multicolumn{3}{@{}c@{\hspace{2em}}}{\textbf{Perfect Match}} & \multicolumn{3}{@{}c@{}}{\textbf{80\% Span Overlap}} \\

& P & R & F1 & P & R & F1 \\ \midrule
ACE05-E & 63.4$^{\pm 2.7}$ & 67.3$^{\pm 2.0}$ &	65.3$^{\pm 2.2}$ &  63.9$^{\pm 2.4}$ & 68.5$^{\pm 1.7}$ & 66.2$^{\pm 1.9}$\\
ACE05-E$^{+}$ & 64.5$^{\pm 1.0}$ & 66.4$^{\pm 2.6}$ & 65.4$^{\pm 1.7}$ & 65.1$^{\pm 0.9}$ & 67.8$^{\pm 2.5}$ & 66.4$^{\pm 1.5}$ \\

ACE05-E$^{++}$ & 57.3$^{\pm 0.8}$ & 59.9$^{\pm 1.2}$ & 58.6$^{\pm 0.9}$ & 63.9$^{\pm 1.1}$ & 66.2$^{\pm 2.1}$ & 65.0$^{\pm 1.6}$ \\

ACE05-E$^{+++}$ & 56.9$^{\pm 0.6}$ & 58.2$^{\pm 0.9}$ & 57.5$^{\pm 0.6}$ & 64.0$^{\pm 0.7}$ & 64.4$^{\pm 1.3}$ & 64.2$^{\pm 0.9}$ \\ \bottomrule
\end{tabular}%
}
\caption{\label{res:args} Results of EventGraph on Arg-C, evaluated with perfect match and token-level span overlap.} 
\label{tab:results2}
\end{table}

\begin{table}[t!]
\resizebox{\columnwidth}{!}{%
\begin{tabular}{rrrr}  \toprule
\textbf{ACE05-E} & \textbf{ACE05-E$^{+}$} & \textbf{ACE05-E$^{++}$} & \textbf{ACE05-E$^{+++}$}  \\ \midrule

88.8$^{\pm 0.4}$ & 87.9$^{\pm 0.5}$ & 96.2$^{\pm 0.2}$ & 96.5$^{\pm 0.6}$ \\ 
 \bottomrule
\end{tabular}%
}
\caption{Results of EventGraph correctly identifying the presence of event(s) in a sentence.}
\label{tab:results3}
\end{table}

\section{Conclusion}

In this paper, we have proposed a new method for event extraction as semantic graph parsing. Our proposed EventGraph has achieved competitive results on ACE2005 for the task of event trigger classification, and obtained new state-of-the-art results for the task of argument role classification. We also provide a graph representation for better visualizing event mentions, and offer an efficient tool to facilitate graph conversion. We create two new datasets from ACE2005, with the full text spans for both triggers and arguments, and offer the corresponding benchmark results. We show that despite adding more and longer text sequences, EventGraph outperforms previous models tested on more restricted datasets. For future work, we would like to experiment with different pretrained language models, and carry out more detailed error analysis. Our codes and models are released as open-source.

\section*{Acknowledgements}

This work was supported by industry partners and the Research Council of Norway with funding to \textit{MediaFutures: Research Centre for Responsible Media Technology and Innovation}, through the centers for Research-based Innovation scheme, project number 309339.

\bibliography{anthology,custom}
\bibliographystyle{acl_natbib}

\clearpage
\newpage

\newpage 

\appendix

\section{Training details}
\label{sec:training-details}

We reuse the training settings from the original PERIN system \citep{samuel-straka-2020-ufal} whenever possible. The model weights are optimized with AdamW \citep{loshchilov2017decoupled} following a warmed-up cosine learning rate schedule. We use a pre-trained multi-lingual XLM\mbox{-}R language model implemented by the HuggingFace \texttt{transformers} library.\footnote{\url{https://huggingface.co/docs/transformers/index}} The hyperparameter configuration is shown in Table \ref{tab:hyperparams}, please consult it with our released code for context: {\small\url{https://github.com/huiling-y/EventGraph}}.

The training was done on a single Nvidia RTX3090 GPU, the runtimes and model sizes (including the fine-tuned language model backbone) for each dataset are given in Table \ref{tab:runtime}. 


\begin{table}[!h]
\resizebox{\columnwidth}{!}{%
\begin{tabular}{@{}lr@{}}
\toprule
\textbf{Hyperparameter} & \textbf{EventGraph} \\ \midrule
batch\_size & 16 \\                    
beta\_2 & 0.98 \\                    
decoder\_learning\_rate & 1.0e-4 \\      
decoder\_weight\_decay & 1.2e-6 \\        
dropout\_transformer & 0.25 \\         
dropout\_transformer\_attention & 0.1 \\
encoder & \textit{"xlm-roberta-large"} \\      
encoder\_learning\_rate & 4.0e-6 \\   
encoder\_weight\_decay & 0.1 \\      
epochs & 180 \\            
hidden\_size\_anchor & 256 \\          
hidden\_size\_edge\_label & 256 \\      
hidden\_size\_edge\_presence & 256 \\   
n\_transformer\_layers & 3 \\                      
query\_length & 2 \\        
warmup\_steps & 1\,000 \\ \bottomrule
\end{tabular}%
}
\caption{Hyperparameter setting for our system, all four datasets use the same configuration.}
\label{tab:hyperparams}
\end{table}

\begin{table}[!h]
\resizebox{\columnwidth}{!}{%
\begin{tabular}{@{}l@{\hspace{3em}}r@{\hspace{2em}}r@{}}
\toprule
\textbf{Dataset} & \textbf{Runtime} & \textbf{Model size} \\ \midrule
ACE05-E & 20:39 h & 341.3  M \\                    
ACE05-E${^+}$ & 21:59 h & 341.3 M \\  
ACE05-E${^{++}}$ & 20:06 h & 341.3 M \\
ACE05-E${^{+++}}$ & 20:03 h & 342.0 M \\ \bottomrule
\end{tabular}%
}
\caption{The training times and model sizes (number of trainable weights) of all our experiments.}
\label{tab:runtime}
\end{table}

\section{Data preprocessing}
\label{sec:preprocessing}

\paragraph{Data splits} All datasets use the same splits\footnote{\url{https://github.com/dwadden/dygiepp/tree/master/scripts/data/ace-event/event-split}} for train/dev/test. Out of the 599 documents, 529 documents are used for training, 30 documents for development, and 40 documents for testing.

\paragraph{ACE-E} We use the preprocessing code\footnote{\url{https://github.com/dwadden/dygiepp}} of \citet{wadden-etal-2019-entity} to obtain the dataset, and they use an older version (v2.0.18) of Spacy\footnote{\url{https://spacy.io/}} for preprocessing.

\paragraph{ACE05-E$^{+}$} We use the preprocessing code\footnote{\url{http://blender.cs.illinois.edu/software/oneie/}} (v0.4.8) of \citet{lin-etal-2020-joint} to obtain the dataset, and they use NLTK\footnote{\url{https://www.nltk.org/}} for preprocessing.

\paragraph{ACE05-E$^{++}$ and ACE05-E$^{+++}$} We use the preprocessing code\footnote{\url{https://github.com/thunlp/HMEAE}} of \citet{wang-etal-2019-hmeae} to obtain the two datasets, and they use Stanford CoreNLP\footnote{\url{https://stanfordnlp.github.io/CoreNLP/}} for preprocessing.



\end{document}